\def\year{2022}\relax
\documentclass[letterpaper]{article} 
\usepackage{aaai22}  
\usepackage{times}  
\usepackage{helvet}  
\usepackage{courier}  
\usepackage[hyphens]{url}  
\usepackage{graphicx} 
\usepackage{booktabs} 
\usepackage{natbib}  
\usepackage{caption} 
\DeclareCaptionStyle{ruled}{labelfont=normalfont,labelsep=colon,strut=off} 
\usepackage{algorithm}
\usepackage{algorithmic}
\usepackage{csquotes}

\usepackage{newfloat}
\usepackage{listings}
\floatstyle{ruled}
\newfloat{listing}{tb}{lst}{}
\floatname{listing}{Listing}
\newcommand{\blank}{$\rule{.5cm}{0.15mm}$ \ }

\title{Commonsense Knowledge Reasoning and Generation \\ with Pre-trained Language Models: A Survey}
\author{
    Prajjwal Bhargava and Vincent Ng
}
\affiliations{
    Human Language Technology Research Institute \\
    University of Texas at Dallas \\
    Richardson, TX 75083-0688 \\
    prajjwal.bhargava@utdallas.edu, vince@hlt.utdallas.edu
}

\begin{document}

\maketitle

\begin{abstract}
While commonsense knowledge acquisition and reasoning has traditionally been a core research topic in the knowledge representation and reasoning community, recent years have seen a surge of interest in the natural language processing community in developing pre-trained models and testing their ability to address a variety of newly designed commonsense knowledge reasoning and generation tasks. This paper presents a survey of these tasks, discusses the strengths and weaknesses of state-of-the-art pre-trained models for commonsense reasoning and generation as revealed by these tasks, and reflects on future research directions.
\end{abstract}

\section{Introduction}
Commonsense knowledge is the information that is generally accepted by the majority of people concerning everyday life, encapsulating the practical knowledge about how the world works. Reasoning with commonsense knowledge is at the core of building natural language understanding models and, more broadly, AI systems that can reason about the world in the same way as humans do.

A vast amount of work on commonsense knowledge acquisition and reasoning has traditionally been conducted in the knowledge representation and reasoning community (see \citet{Zang2013ASO} for a survey). For instance, there have been notable attempts to manually create large-scale commonsense knowledge bases (e.g., Cyc \cite{10.1145/219717.219745} and automatically acquire commonsense knowledge from the Web (e.g., Open Mind Common Sense \cite{Singh2002OpenMC}. More recently, the Winograd Schema Challenge, a pronoun resolution task that requires the use of commonsense knowledge, was proposed as a practical alternative to the Turing Test \cite{10.5555/3031843.3031909}.

The advent of the neural natural language processing (NLP) era has revolutionized virtually all areas of NLP research. One of the major breakthroughs is arguably the development of \emph{pre-trained} language models (PLMs). Specifically, researchers have discovered that neural models can be trained (via a process known as \emph{pre-training}) on a large body of \emph{unannotated} text to acquire general knowledge about language, including both linguistic and commonsense knowledge. This has sparked tremendous interest in the NLP community in examining what kind of commonsense knowledge PLMs possess and the extent to which such knowledge can be exploited to address commonsense knowledge reasoning and generation tasks in the last few years.

Our goal in this paper is to provide the general AI audience with a timely survey of the recent advances in the NLP community on commonsense knowledge reasoning and generation using PLMs. Specifically, the focus of this survey is (1) the kind of commonsense knowledge that PLMs possess and (2) the extent to which such knowledge can be exploited for recently designed commonsense knowledge reasoning and generation tasks. For an overview of what \emph{linguistic} knowledge PLMs possess, we refer the reader to \citet{rogers-etal-2020-primer}. For comprehensive surveys of the details and the inner workings of PLMs, we refer the reader to \citet{DBLP:journals/corr/abs-2003-08271, HAN2021, DBLP:journals/corr/abs-2108-05542}.

\section{Pre-trained Language Models}
\label{c:plm}

In this section, we provide the reader with the background on PLMs needed to understand the rest of the paper.

For a long time, \emph{supervised} learning has been the most successful learning paradigm in NLP. For instance, training a neural model to perform a classification task in a supervised manner primarily involves training an \emph{encoder} to encode the input as a \emph{task-specific representation} that would be useful for classifying a given sample. In contrast, for a text generation task (e.g., text summarization, machine translation), one would typically employ an encoder-decoder neural architecture, in which the encoder first encodes the input, and then the decoder generates the output sequence token by token based on both the encoded input and the tokens that have been generated so far. The performance of supervised models is often limited by the (typically small) amount of data they are trained on.

Pre-training offers a solution to the aforementioned data scarcity problem. The idea is to first train a model on one or more \emph{self-supervised} learning tasks during a process known as \emph{pre-training}, and the resulting model, in which the weights have already been initialized during pre-training, can be \emph{fine-tuned} using the (potentially small amount of) task-specific data in a supervised fashion. Self-supervised learning tasks are NLP tasks for which the label associated with a training instance can be derived automatically from the text itself. Consider, for instance, one of the most well-known self-supervised learning tasks, Masked Language Modeling (MLM) (Devlin et al. 2019). Given a sequence of word tokens in which a certain percentage of tokens is \emph{masked} randomly, the goal of MLM is to predict the masked tokens. As can be imagined, a model for MLM can therefore be trained on instances where each one is composed of a partially masked sequence of word tokens and the associated “class” value is the masked tokens themselves. Because no human annotation is needed, a model can be pre-trained on a very large amount of labeled data that can be automatically generated. Various studies have shown that pre-training allows a model to learn universal language representations that encode both linguistic and commonsense knowledge, and a PLM, after being fine-tuned, can offer substantially improved performance on a wide variety of NLP tasks.

Existing PLMs differ primarily in terms of (1) what is being pre-trained (is it the encoder, the decoder, or both of them?); (2) the self-supervised learning tasks used; and (3) the network architecture. While early work has focused on pre-training the encoder (e.g., BERT \cite{devlin-etal-2019-bert}, RoBERTa \cite{DBLP:journals/corr/abs-1907-11692}, ELECTRA \cite{Clark2020ELECTRA:} or the decoder (e.g., GPT-2 \cite{10.1145/3368089.3417058}), recent work has focused on jointly pre-training the encoder and the decoder (e.g., T5 \cite{2020t5}, BART \cite{lewis-etal-2020-bart}). The most successful PLMs are based on the Transformer \cite{NIPS2017_3f5ee243}, a fully-connected self-attention model. Throughout this paper we will simply use the term PLMs to refer to Transformer-based PLMs.

\section{Capturing Commonsense Knowledge}
\label{c:capture}

How well do PLMs capture commonsense knowledge ? Researchers have employed \emph{probing} to answer this question. To probe a PLM for commonsense knowledge, most of the probing methods use a hand-crafted \emph{template} to convert a relational fact from a knowledge base (KB), which is typically represented in the form of a triple \textless$s,r,o $\textgreater \ where $s$ is the \textsc{Subject}, $r$ is the \textsc{Relation}, and $o$ is the \textsc{Object}, into a natural language sentence. One template needs to be defined for each relation. As illustrated in Figure 1, each triple having the KB relation \enquote{place of birth} would be translated to a sentence of the form \enquote{\textsc{Subject} was born in \textsc{Object}}. Note that a template keeps the entities intact while approximately the \textsc{Relation} to a set of hand-coded verbs/relations that can generalize on numerous entities (e.g., \textsc{Atlocation} may be translated to \enquote{is in}). One of the entities (i.e., the \textsc{Subject} or the \textsc{Object}) in the sentence will then be masked. The resulting \emph{clozed} sentence can then provide an automated and flexible way to probe PLMs for stored knowledge. Specifically, if a PLM can fill in the blank in the given clozed sentence correctly (i.e., the answer is the same as the entity that appears in the triple originally used to generate the sentence), then the PLM is considered to possess the knowledge being probed. Note that we are not asking the PLM to derive \emph{new} knowledge: only inference is performed by the PLM to check for stored knowledge.

Several key observations are being revealed via probing experiments. First, \textbf{PLMs are becoming a promising alternative to KBs}. BERT shows signs of capturing relational knowledge in a zero-shot setting reasonably well compared to supervised alternatives \cite{petroni-etal-2019-language}. It can recall factual knowledge for some relations such as one-to-one, but it struggles to perform well on other relations such as N-to$\mathbf{M}$ and N-to-one.

\begin{figure}[!t]
    \centering
    \includegraphics[width=\linewidth]{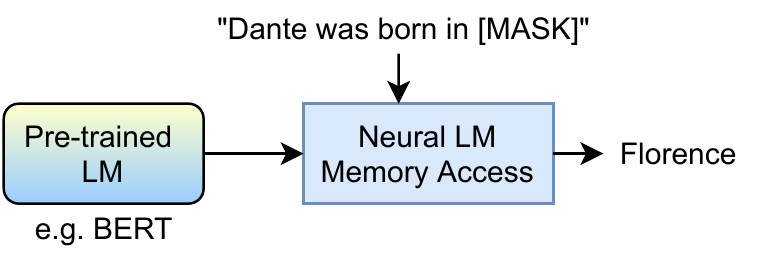}
    \caption{A common approach to probe PLMs for stored knowledge ~\protect\cite{petroni-etal-2019-language}.}
    \label{fig:petroni-kb}
\end{figure}


\begin{table}
\begin{tabular}{
p{.55\columnwidth}p{.3\columnwidth}}
\toprule
\textbf{Prompt} & \textbf{Model Predictions} \\
\midrule
\emph{A \blank has fur, is big, has claws, has teeth, is an animal, eats, is brown, and lives in woods.} & \textbf{bear}, wolf, cat, ... \\
\bottomrule
\end{tabular}
\caption{Masked token predictions about stereotypical assumptions get refined as more properties are appended. ~\protect\cite{Weir-et-al:2020}.}
\label{tab:weir-concept}
\end{table}

Second, \textbf{PLMs do not generalize well on unseen entities}. While BERT can predict the top 100 triples mined from Wikipedia fairly accurately, which suggests that BERT can generalize to specific data sources \cite{davison-etal-2019-commonsense}, PLMs do not generalize well on entities not encountered during pre-training due to their heavy reliance on memorization in the pre-training process \cite{logan-etal-2019-baracks}.

Third, \textbf{PLMs can perform comparisons and categorization of entities}. Specifically, they can compare physical objects along a particular attribute such as weight or size (e.g., a chair is \emph{smaller than} a room) \cite{goel-etal-2019-pre}. When it comes to categorization, they work reasonably well on knowledge types that are ontological in nature, such as \enquote{mango isA fruit} \cite{Hwang2021COMETATOMIC2O}.

Finally, analyzing the top-$k$ predictions made by PLMs on the association between an entity and its attributes, we see that \textbf{PLMs can learn stereotypical associations reasonably well} from large text corpora. As more properties are appended to provide contextual knowledge (see the example in Table 1), the performance of RoBERTa-L increases, with predictions going from being sensible to more acceptable as per human interpretation. Specifically, PLMs do better on functional (e.g., \enquote{eat fish}) and encyclopedic (e.g., \enquote{are found in forests}) knowledge than visual-perceptual variants (e.g., \enquote{bears have fur}). Although this result is encouraging, when asked for widely acceptable properties about an entity, the ranked predictions provided by PLMs do not correlate strongly with those of humans \cite{Weir-et-al:2020}.

\section{Reasoning with Commonsense Knowledge}
\label{c:reasoning}

In this section, we examine how well PLMs perform commonsense reasoning by considering five types of commonsense reasoning tasks.
\subsection{Linguistic Reasoning}
Linguistic reasoning is concerned with understanding text for which the correct interpretation requires commonsense knowledge. A representative benchmark for linguistic reasoning is \textsc{Winogrande} \cite{Sakaguchi2020WINOGRANDEAA}, which consists of Winograd schema-inspired problems that require linguistic, social or physical reasoning \cite{10.5555/3031843.3031909}. As an example, given the sentence \enquote{The plant took up too much room in the urn, because the \rule{0.5cm}{0.15mm} was large} and two answer candidates \enquote{plant} and \enquote{urn}, the goal is to determine which candidate should be used to fill in the blank.

Several observations can be made based on the performance of PLMs on this and other linguistic reasoning tasks.

First, \textbf{BERT shows poor linguistic sensitivity} and becomes fragile on negated and misprimed sentences \cite{kassner-schutze-2020-negated, Ettinger2020WhatBI}. For example, BERT fails to distinguish between the two sentences \enquote{Birds cannot $[\mathrm{MASK}]$ }) and \enquote{Birds can [MASK]} and tends to get distracted if they are prepended with \enquote{misprimes} such as \enquote{Talk? Birds can [MASK]}. The fact that its predictions do not change with such major changes indicates that BERT does not attend to the prominent cues in the desired manner. 

Second, \textbf{PLMs perform poorly on numerical knowledge out-of-the-box} \cite{lin-etal-2020-birds, DBLP:journals/corr/abs-1909-07940, chen-etal-2019-codah, bhagavatula2020abductive}. For instance, given the sentence \enquote{Birds have [MASK] legs}, BERT predicts \enquote{four} to be the answer, suggesting that pre-training does not facilitate the acquisition of numerical knowledge. 

Third, \textbf{as the sentences in a given reasoning task require more turns of logical reasoning (i.e., the task becomes increasingly complex), BERT's performance deteriorates} \cite{DBLP:conf/aaai/ZhouZCH20, richardson-sabharwal-2020-qa, huang-etal-2019-cosmos}. These are typically sentences with complex semantics such as riddles, where PLMs are required to understand figurative language \cite{DBLP:journals/corr/abs-2101-00376}.

Several attempts have been made to improve the \emph{robustness} of PLMs for linguistic reasoning tasks.

\textbf{Semantic similarity}. \citet{niu-etal-2021-semantic} show that semantic similarity matching can be used to make PLMs robust against irrelevant factors such as word frequencies. Specifically, we can first use PLMs to generate plausible answers so that we can compute the similarity between each generated answer and each of the provided answer candidates, and then we can select the answer candidate that has the highest similarity score as the correct answer. \\
\textbf{Attention maps}.  \citet{klein-nabi-2019-attention} show that attention maps obtained from BERT can be used for coreference resolution in long sentences, suggesting their potential usefulness for pronoun disambiguation. \\
\textbf{Numerical reasoning}. To improve numerical reasoning, \citet{geva-etal-2020-injecting}  pre-train BERT on two numerical tasks, one involving predicting what comes after a sentence such as \enquote{3+4+5} and the other involving answering numerical questions (e.g., given a historical passage about Spain, answer the question of \enquote{How many Japanese families were in Spain?}), so that BERT is endowed with the ability to understand computations expressed in pseudo-natural language (text+digits).

\begin{figure}[!t]
    \centering
    \includegraphics[width=0.9\linewidth]{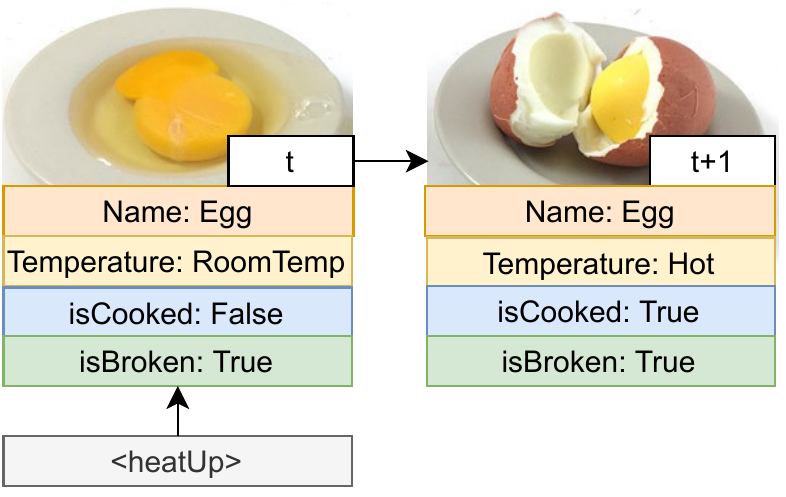}
    \caption{Simulating what might happen next in order to enable LMs to encode language \enquote{form} and \enquote{meaning} ~\protect\cite{zellers2021piglet}.}
    \label{fig:piglet}
\end{figure}

\subsection{Reasoning about Physical World}
Physical commonsense reasoning involves understanding concepts based on the physical properties of objects, including the \emph{affordances} of objects (i.e., the actions applicable to them) and how they can be manipulated. A representative benchmark for physical commonsense reasoning is PIQA \cite{Bisk2020}. Given a sentence such as \enquote{When boiling butter, when it's ready, you can \blank} , the goal is to fill in the blank with one of two answer options, such as \enquote{Pour it onto a plate} and \enquote{Pour it onto a jar}.

Perception and Interaction are among the key components behind how humans learn to reason about the physical world. However, static input representations which current neural models are fed are inadequate since they cannot compensate for the information humans derive from being in a dynamic physical world. So, a key question posed by PIQA is whether PLMs can reason over physical commonsense questions without interacting with the physical world.

Several observations can be made about the performance of PLMs on physical commonsense reasoning questions. First, \textbf{PLMs predominantly learn property associations that are explicitly mentioned in text}, achieving higher accuracies on entities that have simple affordances (e.g., \enquote{spoon}) than on entities that have a a long tail of affordances (e.g., \enquote{water}). Second, \textbf{PLMs struggle to understand fundamental relations} (e.g., \enquote{before/after}, \enquote{top/bottom}, and \textbf{find it hard to reason when common objects are used in unconventional ways} (e.g., a glue stick is used as a paper weight). Finally, although neural representations are dexterous at capturing the affordances (\enquote{boats can be driven}) and properties (\enquote{boats require fuel}) of objects, \textbf{PLMs struggle to understand the connection between affordances and properties} \cite{Forbes2019DoNL, zhao-etal-2020-learning}. \\

In light of the weakness associated with the lack of interaction with the physical world, \cite{zellers2021piglet} explore the benefits of providing PLMs access to world dynamics. World dynamics include information that one would obtain after interacting with objects. As an example, consider Figure 2. If an action \texttt{heatUp} is applied to a pan, the model will learn that the temperature of an egg has risen to become \texttt{hot} and is now in a \texttt{cooked} state. Predicting object states after an action has been taken drastically improves a PLM's ability to make correct inferences about object states.

\subsection{Abductive Reasoning}
Abductive reasoning involves finding the most likely explanation for a set of incomplete observations. There are at least two representative benchmarks for abductive reasoning, \textsc{CosmosQA} \cite{huang-etal-2019-cosmos} and \textsc{Hellaswag} \cite{zellers-etal-2019-hellaswag}. \textsc{CosmosQA} is a commonsense comprehension task where, given a context, the goal is to choose the answer to the question based on the context from four answer candidates. This benchmark contains questions that require abductive reasoning, such as \enquote{what might I continue to do after the situation described in the context?} \textsc{Hellaswag} is a benchmark in which the goal is to choose the best plausible ending of a given context out of four options.

Several observations can be made. First, \cite{huang-etal-2019-cosmos} attribute the errors made by their model on \textsc{Cosmosqa} to two reasons. First, \textbf{PLMs struggle on examples where the context becomes intricate enough to require cross-sentence interpretation and reasoning}. In such examples, PLMs are required to understand the important parts of the passage and jointly attend to the identified parts. In addition, \textbf{PLMs do not understand what situations are inconsistent with human commonsense}. For example, they may choose \enquote{leaving a baby alone at home is not safe} over \enquote{she would try to find a babysitter} when asked the question \enquote{what would happen if she does not find a daycare}. Interestingly,when one of the answer options is \enquote{None of the above}, PLMs often struggle to choose this option since the words in this option do not provide enough signal for why this option might be correct.

Second, \textbf{PLMs struggle with selecting the most plausible ending given a context} for \textsc{Hellaswag}. Since a given context can have multiple correct endings, determining which one would be the most plausible requires prior reasoning of what humans relate to the most with their commonsense knowledge. \citet{zellers-etal-2018-swag, zellers-etal-2019-hellaswag} show that when more surface cues are eliminated, PLMs are less likely to be able to predict the most plausible ending even though it might be trivial for humans to do so.

\begin{figure*}[!t]
    \centering
    \includegraphics[width=0.9\linewidth]{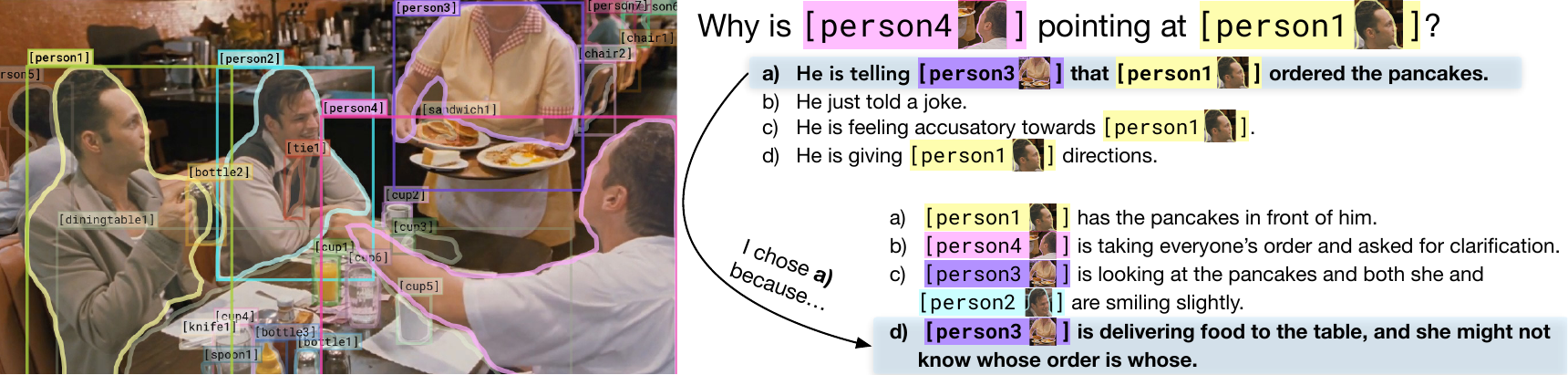}
    \caption{Learning to reason about dynamic context from a static image ~\protect\cite{zellers2019vcr}.}
    \label{fig:vcr}
\end{figure*}

\subsection{Social Reasoning}
Social reasoning involves modeling the mental states of others and their likely actions to the extent that reasoning can be performed over their behaviors and emotions. A representative benchmark for social reasoning is \textsc{Socialiqa} \cite{Sap2019SocialIC}, which evaluates commonsense reasoning based on social situations and interactions. Consider the following example taken from \textsc{socialiqa}, in which the correct answer is boldfaced: \\

\enquote{Context}: \enquote{Tracy had accidentally pressed upon Austin in the small elevator and it was awkward.} \\
\enquote{Question} : \enquote{Why did Tracy do this?} \\
\enquote{Choice A}: \enquote{get very close to Austin}; \\
\textbf{\enquote{Choice B}: \enquote{squeeze into the elevator}}; \\
\enquote{Choice C}: \enquote{get flirty with Austin} \\

Several observations can be made. First, BERT finds examples of effects (\enquote{what will happen to X?}) and motivation (\enquote{why did X do that to Y?}) easier than those that involve understanding descriptions (\enquote{how would you describe X?}) \cite{Sap2019SocialIC}. Second, it performs better on examples where the answer exhibits cues about emotions than those involving spatial commonsense \cite{bhagavatula2020abductive}. 

\subsection{Multimodal Reasoning}
Textual representations are restricted to what can be expressed through natural language and therefore are unable to represent the multi-modal information that humans could have access to or infer from by being in a dynamic world, such as a constant stream of images and a sequence of interactions in the physical world. Vision naturally becomes a next step towards enabling learning through joint interaction (Baldwin 1995). However, merely using raw visual images along with their descriptions is by no means sufficient to provide grounded understanding \cite{marasovic-etal-2020-natural}. For example, inferring the intentions of the entities in images can only be well dealt with if we have some prior information (either behavioral or temporal) to rely on to make justifiable inferences. This has led the community to look into approaches that can help provide a tighter integration of linguistic and visual modalities.

There are two well-known benchmarks for multimodal reasoning. \textsc{Visual commonsense reasoning (Vcr)} \cite{zellers2019vcr} seeks to answer cognition-level questions from images. Concretely, given an image with a list of regions and a question, the goal is to choose the answer to the question out of a set of possible candidates and provide a rationale that can explain why the chosen answer is correct. An example can be found in Figure 3. \textsc{Visual commonsense graphs} \cite{park2020visualcomet} checks how well PLMs reason about the dynamic context from a static image and an event. Specifically, given an image and a textual description of an event at present, the task is to generate the rest of the visual commonsense graph that is connected to the event. For example, given an image of a man who is drowning in the river and a textual description of the event, the goal is to generate a commonsense graph with nodes such as \enquote{the man wanted to save himself from drowning}, \enquote{the man is waiting for help}, \enquote{the man senses his own death}, and \enquote{the man needs to swim towards the river bank}. Empirical results reveal that for both benchmarks, models that exploit both visual and textual information outperform those that only use textual information. This suggests that \textbf{visual features help make higher quality commonsense inferences}.

\subsection{Temporal Reasoning}
Time is an inherent aspect of events. Broadly, temporal reasoning involves two subtasks. \emph{Temporal attribute prediction} involves understanding an event mentioned in text or dialogue through reasoning with its temporal dimensions such as the duration of the event, when the event typically happens, how long the event is going to be stationary, and how often it happens. \emph{Temporal relation identification} involves understanding how an event is temporally related to other events mentioned in the same text or dialogue (e.g., did an event take place \emph{before} or \emph{after} another event?). Temporal reasoning is challenging because the timestamp associated with an event and the aforementioned temporal dimensions may not be mentioned explicitly and therefore need to be inferred.

Two commonly-used benchmarks have been developed for temporal reasoning. \textsc{Mc-taco} \cite{zhou-etal-2020-temporal} is a question-answering benchmark involving temporal commonsense comprehension. Here is an example: \\

\enquote{Context}: The massive ice sheet, called a glacier, caused the features on the land you see today \\
\enquote{Question}: When did the glacier start to impact the land's features ? \\
\enquote{options}: \textbf{a) centuries ago}; b) hours ago; c) 10 years ago; \textbf{d) tens of millions of years ago} \\

\textsc{Timedial} \cite{qin-etal-2021-timedial} involves temporal reasoning in dialogues. Here is an example: \\
A: May we see the wine list please. \\
B: Sure. Our special wine today is a 1989 Chardonnay. \\
A: I'd like a bottle please. \\
B: I'll need to see your ID please. \\
A: Here you go. \\
B: Sorry about the inconvenience, you look so young. I had to make sure you are over. \\
\textbf{a) 21 years old}; b) 30 years old; c) 4 years old; \textbf{d) 18 years old} \\

Ideally, one can train \emph{time-aware} PLMs to address these temporal reasoning tasks. An obstacle to the development of such PLMs concerns the lack of large-scale KBs that incorporate the notion of time into the facts that they encode over entities and events. For instance, the \textsc{Location} relation (i.e., where a person lives) and the \textsc{Employment} relation (i.e., the company a person is affiliated with) are dependent on time, but existing KBs typically fail to encode the time period for which a given relation holds true. Such time-aware KBs should also encode temporal commonsense knowledge such as \enquote{if a student attends a university, s/he will likely graduate and work after a few years}.

Given the lack of such KBs, time-aware PLMs can only be trained on the annotated training data provided by \textsc{Mctaco} and \textsc{Timedial}. For instance, \cite{zhou-etal-2020-temporal} propose \textsc{Taco-lm}, a \textsc{Bert}-based PLM that is trained to be temporally aware by contextually estimating duration and time via (1) extracting the important events and their temporal information, and then (2) asking the model to predict the masked tokens that talk about some temporal aspect. However, \textsc{Taco-lm} only provides marginal improvements over \textsc{Bert} w.r.t. duration, frequency, and when the event typically takes place. More recently, \citet{qin-etal-2021-timedial} have shown that fine-tuned PLMs struggle to perform well on \textsc{Timedial} primarily because they largely fail to understand the context of the given dialogue and instead simply rely on shallow cues about the temporal patterns in the context.

\section{Generating Commonsense Knowledge}
\label{c:generate}
Commonsense knowledge generation is a critical component in building commonsense knowledge resources. Broadly, we can divide commonsense knowledge generation tasks into two categories, as described below. 

\subsection{Knowledge Base Completion}
A KB is a collection of relational facts, each of which is represented as a triple \textless$s, r, o$\textgreater, where $s$ is the \textsc{subject}, $r$ is the \textsc{relation}, and $o$ is the \textsc{object}. KB completion is the task of automatically inferring missing facts by reasoning about the information already present in the KB.

To date, the most successful knowledge generation approach with PLMs is arguably Commonsense Transformer (\textsc{COMET}) \cite{Bosselut2019COMETCT}. \textsc{COMET} can be used to generate $o$ given $s$ and $r$, after being pre-trained on a knowledge base such as ConceptNet \cite{Singh2002OpenMC, DBLP:journals/corr/SpeerCH16}, which represents (mostly taxonomic) commonsense knowledge as a graph of concepts (words or phrases) connected by relations (edge types), or \textsc{ATOMIC} \cite{DBLP:conf/aaai/SapBABLRRSC19}, which is a large-scale KG consisting of textual descriptions of inferential knowledge (\emph{if-then} relations).

\subsection{Constrained Commonsense Text Generation}
Next we examine studies on how PLMs can be used to generate commonsense text subject to a set of constraints. \\

\textbf{Tasks} There are three benchmarks commonly used to evaluate commonsense generation approaches.

\textsc{Commongen} \cite{lin-etal-2020-commongen}: Given a concept set (e.g., $\{$\texttt{dog, frisbee, catch, throw}$\}$), the goal is to generate a coherent sentence describing an everyday event using all the provided concepts.

\textsc{Commonsense explanations} (COS-E) \cite{rajani-etal-2019-explain}: is a dataset targeted at a task known as explanation generation. Given that a model selects an answer (from a set of candidates) to a given question, the goal is to generate an explanation of why the selected answer is correct. The resulting explanation may help us understand the reasoning that a model relies on to arrive at the selected answer.

$\alpha$ NLG \cite{bhagavatula2020abductive}: Given two observations/events $O_{1}$ and $O_{2}$ that happen in two different timesteps, the goal is to generate a valid hypothesis $h$ of what happened between the observations/events. \\

\textbf{Challenges} These benchmarks reveal that PLMs, when used as commonsense knowledge generators, suffer from several shortcomings. 
\begin{itemize}
    \item \textbf{Poor coherency}: The generated sentences do not necessarily adhere to the human notion of commonsense. For instance, given the concept set \{dog, catch, throw, frisbee\}, GPT2 generates the sentence \enquote{A dog throws a frisbee at a football player}. Although this sentence is grammatically correct, it suffers from poor coherency.
    \item \textbf{Insufficient concept coverage}: PLMs continue to produce sentences that fail to include all concepts from the provided concept set. In the previous example, the concept \enquote{catch} was not used to generate the output.
    \item \textbf{Limited reasoning capability}: It is not clear what kind of reasoning is used by PLMs to arrive at an answer. Though a solution to explanation generation could shed light on this question, some studies show that existing approaches generate either trivial or noisy explanations, providing little or no evidence of how a PLM arrives at the selected answer  \cite{ji-etal-2020-generating}. Other studies show that the reasoning used by PLMs is often not fully correct \cite{mccoy-etal-2019-right, shwartz-choi-2020-neural}. Overall, the reasoning capability of PLMs is far from satisfactory.
\end{itemize}

\begin{figure}[!t]
    \centering
    \includegraphics[width=0.9\linewidth]{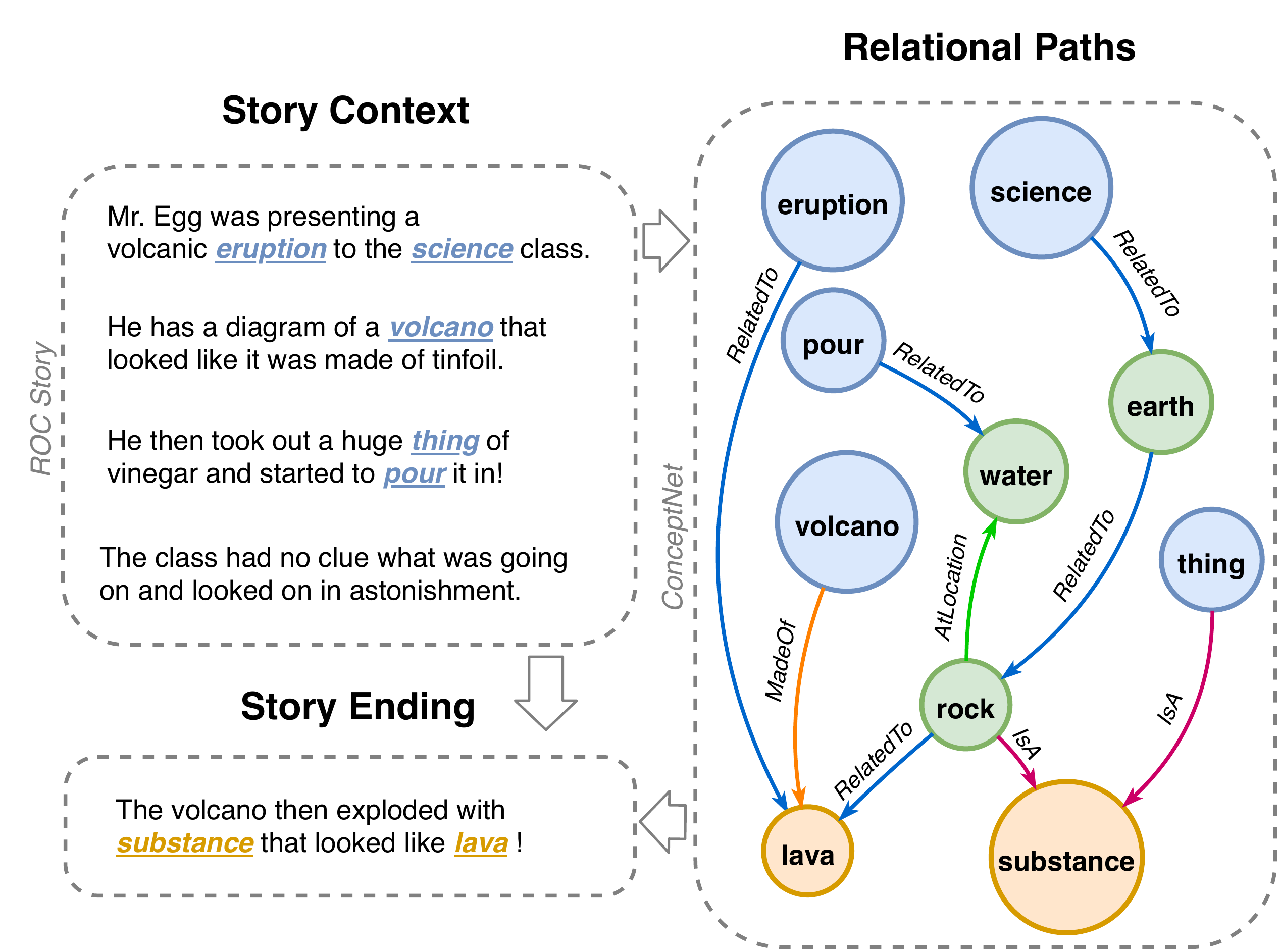}
    \caption{Using structural relational knowledge for Multi hop reasoning ~\protect\cite{ji-etal-2020-language}. Blue nodes correspond to concepts in story, orange nodes correspond to those in story ending and green nodes are intermediate concepts that connect their blue and orange counterparts.}
    \label{fig:multi_hop}
\end{figure}

\textbf{Improving Coherency and Concept Coverage} Several approaches have been proposed to address two of the challenges, poor coherency and insufficient concept coverage. \\
\textbf{Using prototypes.} \cite{guu-etal-2018-generating} propose a sentence generation mechanism that involves selecting a sentence from the training data (known as a \emph{prototype}) and editing it into a form that satisfies a given set of constraints. Their hypothesis is that it is easier to edit a sentence that is grammatically correct and semantically coherent than to generating one from scratch. If provided with a concept set \{\texttt{trailer, shirt, side, sit, road}\} from \textsc{commongen}, a PLM may generate \enquote{A man sits on the side of a trailer and a shirt}, whereas a prototype such as \enquote{Two guys in red shirts are sitting on chairs, by the side of the road, behind that open trailer} maybe edited by the PLM to form \enquote{a man in a white shirt and black pants sits on the side of a trailer on the road}, which has better coverage and coherency. \\
\textbf{Using knowledge graphs (KGs).} KGs can play a crucial role in enabling PLMs to improve the semantic correctness of text (and thus coherency) of text as they can provide PLMs with information that may not be captured reliably by PLMs,  such as entity representations and their dependency relations (i.e., how concepts are related), which may not be captured reliably by PLMs. For instance, Li et al. (2020) extract concept-specific relations from a KG and inject them into a PLM to make the generated text more coherent. \\
\textbf{Reasoning over multi-hop relational paths in KGs}. The sparse connections between the nodes in KGs may make it hard for PLMs to learn rich relations from them. These rich relations, however, may be needed by PLMs to generate commonsense sentences with rich semantic structures in order to boost coherency. A solution to this problem is to perform \emph{multi-hop reasoning}, which involves reasoning over multiple edges/relations in a KG. When performing multi-hop reasoning, models are required to attend to different parts of a given context to answer a question. Figure 4 shows an example of a task known as \emph{story ending} generation, where the goal is to generate the end of a story given the story context. In this example, external commonsense knowledge in the form of relational paths can guide the generation of the two nodes \texttt{lava} and \texttt{substance} by providing background knowledge such as (volcano, madeOf, lava). Capturing what \texttt{lava} and \texttt{substance} that appear in the story ending refer to in the story context is a non-trivial task for PLMs, especially when the story is long. To address this drawback, \cite{ji-etal-2020-language} perform reasoning over multi-hop relational paths as a way to extract structural and semantic knowledge from a KG.\\
\textbf{Using iterative refinements}. When provided with information about past and future events (as in $\alpha \mathrm{NLG}$ ), humans can easily reason about these events by using contextual and prior knowledge. This kind of non-monotonic reasoning is crucial to improving generation coherency. However, nonmonotonic reasoning is difficult for neural models since the generation process happens predominantly while conditioning on the left context \cite{Welleck2019NonMonotonicST}. To address this problem, \cite{qin-etal-2020-back} propose a decoding approach that involves sampling from the combined output vector representations computed from both forward and backward propagation. In other words, iterative refinements are made on the generated text through alternating between forward and backward passes, yielding generated text with improved coherency.

\section{Concluding Remarks}
\label{c:conclusion}

Despite recent progresses on using PLMs to address commonsense knowledge reasoning and generation tasks, many of these tasks are far from being solved. We conclude our discussion with key challenges in this area of research.

\addtolength{\tabcolsep}{-4pt} 
\begin{table}[h]
    \small
    \centering
    \begin{tabular}{llllll}
        \toprule
        Dataset             & Model & \multicolumn{1}{c|}{Human} & Dataset             & Model & Human \\
        \hline
        \textsc{hellaswag}  & 93.85 & \multicolumn{1}{c|}{95.6}  & \textsc{winogrande} & 86.64 & 94.0 \\
        \textsc{comosqa}    & 91.79 & \multicolumn{1}{c|}{94.0}  & \textsc{socialiqa}  & 83.15 & 88.1 \\
        \textsc{piqa}       & 90.13 & \multicolumn{1}{c|}{94.9}  & \textsc{vcr}        & 63.15.& 85.0  \\
        \bottomrule
    \end{tabular}
    \label{tab:sota}
    \caption{Results of state-of-the-art models and human baselines on widely-used commonsense reasoning benchmarks.}
\end{table}

\textbf{Improving benchmarks.} While state-of-the-art models have achieved near-human performance on many of the benchmarks mentioned in this paper (see Table 2), the reasoning tasks underlying these benchmarks are still far from being solved. Consequently, it is not clear what the performance gains on a particular benchmark mean. \citet{bender-koller-2020-climbing} point out that acing a benchmark has led us to over-estimate the capability of PLMs, which in turn has given rise to misleading definitions of \enquote{understanding}. It is therefore important to re-think what is being learned by PLMs and how benchmarks can be made more representative such that performance gains on them translate to meaningful progress towards the bigger goal of \enquote{understanding}. \\
\textbf{Reducing biases.} Biases in benchmarks such as predictable question structures, annotation artifacts, and lexical overlap provide easy shortcuts for PLMs to arrive at correct answers without involving reasoning. To mitigate biases, researchers have used \emph{adversarial filtering} wherein easily solvable options are replaced iteratively by new ones until the discriminator misclassifies it \cite{zellers-etal-2018-swag, mccoy-etal-2019-right, DBLP:conf/icml/BrasSBZPSC20}. To robustify data, several workarounds have been proposed that revolve around reducing lexical overlap, creating complex reasoning questions that require additional context, and employing adversarial approaches with newer models \cite{gardner-etal-2019-making}. Bias reduction in benchmarks remains an active research area.\\
\textbf{Exploring new components of commonsense knowledge.} These are numerous components of commonsense knowledge that are partially understood and not covered by the present research. One primary reason for this is that we do not have a comprehensive understanding of how humans learn. A concrete example can be derived from Kahneman's (2011) cognitive system of intuition. There is no clear way of representing a human's mental and emotional states that can be readily used by our algorithms. Modeling multiple mental states with natural language is a highly non-trivial process \cite{sap-etal-2020-commonsense}. \\
\textbf{Addressing the reporting bias.} Much commonsense knowledge is assumed rather than mentioned explicitly in text \cite{Grice1975-GRILA-2, Durme2009ExtractingIK, 10.1145/2509558.2509563}. This results in what is known as the \enquote{reporting bias}, which, when combined with the scarcity of training data for many NLP tasks, makes it hard for PLMs to receive appropriate signal about a particular concept \cite{zhao-etal-2020-learning} . This could lead to over-generalization of associations and amplification of biases \cite{shwartz-choi-2020-neural}. How to address the reporting bias remains an open question. \\
\textbf{Improving existing KGs}. While many approaches rely on KGs to obtain rich contextual knowledge, existing KGs have at least two key weaknesses that can potentially limit their usefulness for commonsense reasoning. The first is \emph{sparsity}: many concepts and relations are missing in existing KGs \cite{li-etal-2016-commonsense}. This sparsity problem in turn limits the amount of knowledge that can be extracted from KGs for commonsense reasoning \cite{zhao-etal-2020-learning}. The second is \emph{non-contextualization}: finding the nodes that are most relevant to a query is difficult, particularly by propagation-based algorithms, because many of them are non-contextual in nature \cite{DBLP:journals/corr/abs-1911-02085}. To address the non-contextualization problem, \cite{malaviya2020commonsense} have attempted to use the structural and semantic connections of the nodes in a KG to obtain contextual information, which addresses the non-contextualization problem. The resulting contextual information can then be explicitly encoded in a KG by creating additional nodes, which alleviates the sparsity problem. How to densify KGs and contextualize their nodes is an ongoing research topic \cite{wang-etal-2020-connecting}. \\
\textbf{Harnessing commonsense knowledge from different modalities}. There are many instances wherein the visual modality is required along with text to make sense of a particular situation \cite{park2020visualcomet}. While humans learn commonsense knowledge predominantly through perception and interaction with the physical world, neural models are primarily trained on text data. Harnessing commonsense knowledge from different modalities can potentially take these models to the next level of performance. \\
\textbf{Towards multilinguality}. An important but unexplored direction is multi-lingual commonsense reasoning and generation. Studies have shown that the performance of cross-lingual PLMs is poor when evaluated on non-English commonsense reasoning benchmarks \cite{DBLP:journals/corr/abs-2106-06937}. These models perform tremendously poorly when evaluated on a test set that was translated to English, leading to staggering transfer reasoning capabilities to other languages and restricting the research scope to only certain languages \cite{ponti-etal-2020-xcopa}.

\bibliography{aaai22}

\end{document}